\documentclass[letterpaper]{article} 
\usepackage[preprint]{aaai2027}
\usepackage[hyphens]{url}  
\usepackage{graphicx} 
\usepackage{natbib}  
\usepackage{caption} 
\usepackage{algorithm}
\usepackage{algorithmic}

\usepackage{newfloat}
\usepackage{listings}
\DeclareCaptionStyle{ruled}{labelfont=normalfont,labelsep=colon,strut=off} 
\floatstyle{ruled}
\newfloat{listing}{tb}{lst}{}
\floatname{listing}{Listing}

\usepackage{booktabs}

\usepackage{amssymb}
\usepackage{amsmath}
\usepackage{multirow}
\usepackage{kotex}
\usepackage{siunitx}
\usepackage{diagbox}
\usepackage{tikz}

\title{Visual Information-Guided Parallel Decoding for \\ Diffusion Multimodal Large Language Models}

\author{
Insu Lee\textsuperscript{\rm 1}\equalcontrib,
Wooje Park\textsuperscript{\rm 1}\equalcontrib,
Wonseok Shin\textsuperscript{\rm 1},
Jinwoo Son\textsuperscript{\rm 1},
Byonghyo Shim\textsuperscript{\rm 1}
}

\affiliations{
\textsuperscript{\rm 1}Seoul National University\\
\{islee, wjpark, wsshin, jinwooson, bshim\}@islab.snu.ac.kr
}

\begin{document}

\maketitle

\begin{abstract}
Diffusion multimodal large language models (dMLLMs) have recently emerged as a new decoding paradigm for multimodal generation. 
Starting from a fully masked sequence, dMLLMs progressively decode the sequence by unmasking a subset of the remaining masked positions at each step.
Since the selected tokens serve as the prediction context for subsequent steps, deciding which tokens to decode is crucial to the quality of the final output. 
The most common strategy prioritizes tokens based on a certainty measure that tends to favor tokens frequently observed in the training data.
Recent approaches instead order tokens according to their influence on subsequent predictions, but do not explicitly account for the input image.
We propose the Visual Information-Guided Sampler (VIG-Sampler), which prioritizes tokens based on their attention to image tokens.
We further impose a constraint that penalizes candidate tokens whose image-attention distributions are similar to those of previously selected tokens, thereby increasing the information gain of the decoded subset.
Extensive experiments on 7 captioning and VQA benchmarks with 3 open-source dMLLMs demonstrate the effectiveness of VIG-Sampler, which outperforms the Info-Gain Sampler by an average of 19.3 CIDEr points across the captioning benchmarks and surpasses it on COCO Caption while using only half as many decoding steps.
\end{abstract}

\section{Introduction}
\begin{figure*}[t!]
  \centering
  \includegraphics[width=\textwidth]{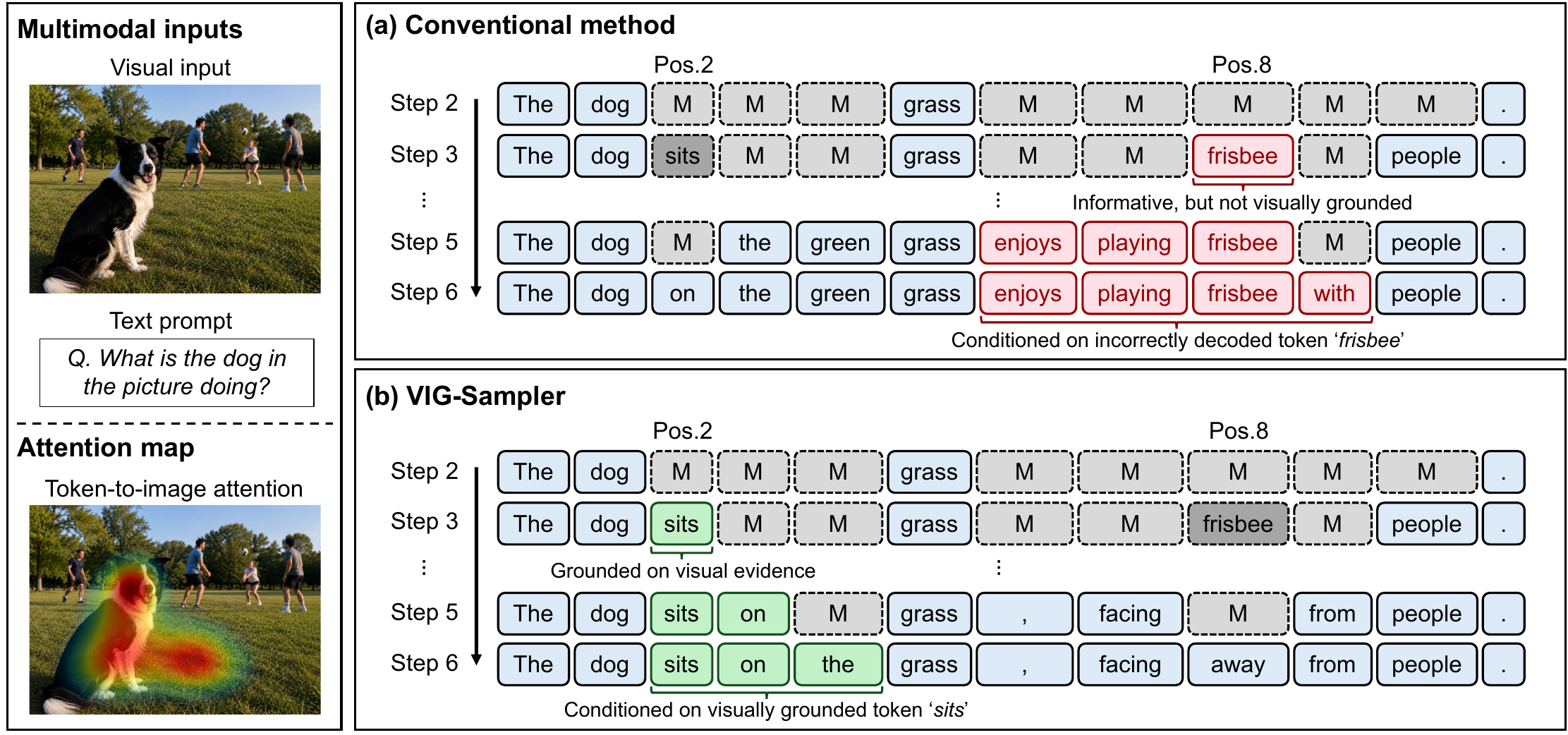}
    \caption{
    Comparison of conventional sampling and VIG-Sampler.
    The conventional method may prioritize the visually unsupported token \textit{``frisbee''}, causing subsequent predictions to drift from the image.
    In contrast, VIG-Sampler uses image attention to prioritize the visually grounded token \textit{``sits''}, leading to a response consistent with the visual input.
    }
  \label{fig:intro}
  \vspace{-4pt}
\end{figure*}

Recent advances in diffusion large language models (dLLMs) have led to the emergence of diffusion multimodal large language models (dMLLMs)~\cite{li2026lavida, llada-v, yang2026mmada}.
By extending the diffusion-based text generation paradigm to multimodal inputs, dMLLMs provide a promising alternative to the conventional \textit{autoregressive} multimodal large language models (AR-MLLMs).
Unlike AR-MLLMs generating a sequence by decoding tokens from left to right, dMLLMs generate a sequence through an iterative decoding process without following a fixed positional order~\cite{kim2025train}.
In this decoding process, the choice of which tokens to decode is of great importance, as the resulting tokens become part of the conditioning context for the subsequent prediction and ultimately affect the quality of the final output~\cite{gwak2025reward, jazbec2025learning}.

The commonly used strategy for this token selection is to rank tokens according to certainty measures derived from the model's predictions and then process them in the ordered sequence~\cite{wu2025fast, llada, ebsampler}. 
One well-known drawback is that this certainty-based selection does not account for the usefulness of the decoded tokens in predicting the remaining masked tokens~\cite{fu2025bits}. 
Since models tend to assign high confidence to tokens frequently observed in the training data, semantically uninformative tokens such as copulas, punctuation marks, and end-of-text tokens are often decoded first in the early steps~\cite{martinez2024mitigating,huang2026empirical}. 
In such cases, these tokens fix the syntactic structure or ending of the response before the model determines what content to generate.
Thus, subsequent predictions are constrained to follow the prematurely determined structure, potentially generating a structurally consistent but incorrect response~\cite{pc-sampler, info-gain}.

To overcome the shortcoming, several token selection approaches that account for the influence of decoded tokens on subsequent predictions have been proposed.
One approach uses token frequencies in a fixed reference corpus to mitigate the excessive prevalence of less informative tokens in generated sequences~\cite{pc-sampler}.
Another approach selects tokens based on each token’s contribution to reducing sequence-level uncertainty (i.e., information gain).
In this scheme, the contribution is quantified by the decrease in entropy over the remaining masked tokens after the token is decoded~\cite{info-gain}.
Although these approaches have improved the generation quality in language-only settings to some extent, they do not explicitly account for the input image and thus generate tokens without sufficient visual grounding.

To illustrate this scenario, consider a dMLLM is given an image of a sitting dog along with the question, \textit{“What is the dog in the picture doing?”} (see Figure~\ref{fig:intro}).
At decoding step 3, the model may predict \textit{“sits”} at position 2 and \textit{“frisbee”} at position 8.
Conventional token selection approaches may decode \textit{“frisbee”} instead of \textit{“sits”} if decoding the token at position 8 is estimated to yield greater information gain, even though the token is not grounded in the visual content.
Once decoded, \textit{“frisbee”} becomes a part of the context for subsequent predictions and may increase the likelihood of tokens such as \textit{“playing”} and \textit{“with”}.
The model can thus generate a response inconsistent with the image, such as \textit{“The dog on the green grass enjoys playing frisbee with people.”}
This example highlights a central principle of dMLLM decoding: visual evidence and language-based signals should be considered jointly.

An aim of this paper is to propose a novel image-grounded sampling strategy for parallel decoding in dMLLMs to select token sets that are both visually grounded and informative for subsequent predictions.
To this end, we propose the Visual Information-Guided Sampler (VIG-Sampler), which leverages image attention readily available at each decoding step.
Based on the observation that attention to image tokens serves as a reliable proxy for both visual grounding and informativeness, VIG-Sampler employs each token's image attention as an ordering score.
This prioritizes visually informative tokens and thereby promotes an image-grounded decoding trajectory.
In addition to the token-level ordering, VIG-Sampler increases the overall information gain from decoding multiple tokens in parallel. 
We empirically observe that set-level information gain can fall substantially below the sum of individual gains when selected tokens convey redundant information, and that this redundancy manifests as similarity between their token-to-image attention distributions.
VIG-Sampler further translates this similarity into a penalty score, discouraging tokens with attention distributions similar to those of the tokens already selected.

Overall, VIG-Sampler jointly promotes visual grounding and complementary information gain during parallel decoding, without requiring additional training or model forward passes.
Extensive experiments on 7 captioning and visual question answering (VQA) benchmarks with 3 open-source dMLLMs show that VIG-Sampler consistently outperforms recent samplers that rely solely on the textual context, regardless of the number of tokens decoded per step.
In particular, when decoding 8 tokens per step with LaViDa~\cite{li2026lavida}, VIG-Sampler outperforms the Info-Gain Sampler by an average of 19.3 CIDEr points on captioning benchmarks and 7.3 accuracy points on VQA benchmarks.
Notably, on COCO captioning~\cite{lin2014microsoft}, VIG-Sampler decoding 8 tokens per step still surpasses the Info-Gain Sampler decoding only 4 tokens per step by 5.3 CIDEr points, despite using half as many decoding steps.
These results demonstrate the effectiveness of VIG-Sampler and show that grounding token selection in visual information is essential for dMLLMs.

\noindent In summary, our contributions are as follows:

\begin{itemize}
\item Building on our analytical findings, we develop VIG-Sampler, a decoding strategy that leverages readily available token-to-image attention to select visually grounded and informative tokens without incurring additional training or model forward passes.

\item Through empirical analysis, we show that tokens decoded in parallel may provide redundant information, limiting the information gain of the selected set, and incorporate this observation into VIG-Sampler through attention-guided subset selection.

\item Extensive experiments on 7 multimodal benchmarks with 3 open-source dMLLMs demonstrate that VIG-Sampler outperforms Info-Gain Sampler by 19.3 CIDEr points on image captioning and still achieves better performance when using only half the number of decoding steps.

\end{itemize}
\section{Visual Information-Guided Sampler}
\label{sec:method}
\begin{figure}[t!]
  \centering
    \includegraphics[width=\linewidth]{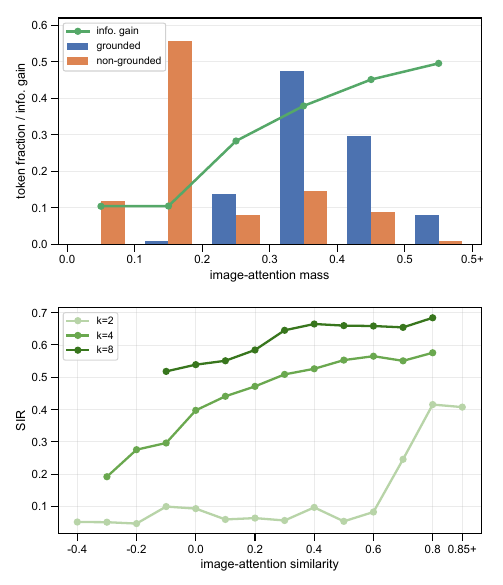}
\caption{Motivating observations on how image attention relates to visual grounding and information gain during dMLLM decoding.
Top: distributions of visually grounded and non-grounded tokens across image-attention-mass ranges, together with the mean information gain.
Bottom: shared information ratio (SIR) versus mean pairwise image-attention similarity for selected token sets of size $k\in\{2,4,8\}$.}
  \label{fig:motivation}
\end{figure}

\begin{figure*}[t!]
  \centering
  \includegraphics[width=\textwidth]{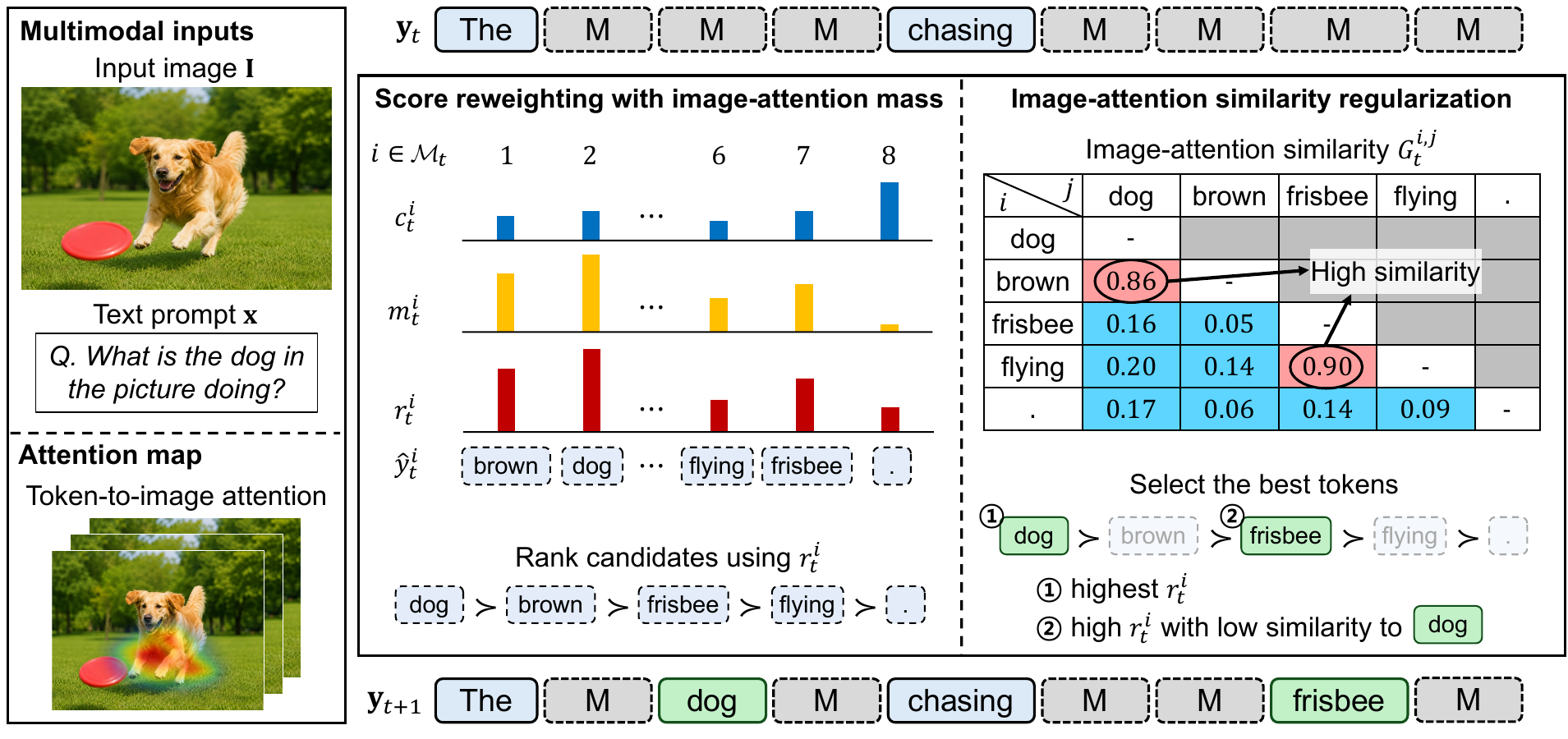}
\caption{
Overview of the proposed VIG-Sampler.
At each decoding step, image-attention mass is used to adjust the confidence score of each masked position, prioritizing tokens that are both visually grounded and informative.
The resulting scores and pairwise image-attention similarities are then used to select the token set expected to yield higher information gain.
}
\label{fig:method}
\end{figure*}

\subsection{Preliminaries}
\label{subsec:preliminaries}

\paragraph{Decoding process of dMLLM.}
We consider a dMLLM $p_\theta$ that generates a textual response $\mathbf{y}$ conditioned on a visual input $\mathbf{I}$ and a text prompt $\mathbf{x}$.
Starting from a length-$N$ fully masked sequence, $\mathbf{y}_0=[\mathrm{MASK}]^N$, we perform $T$ decoding steps to produce the final response $\mathbf{y}_T=\mathbf{y}$.
At step $t$, the model predicts a distribution $p_\theta^i(\cdot \mid \mathbf{I}, \mathbf{x}, \mathbf{y}_t)$ for each masked position $i \in \mathcal{M}_t$, where $\mathcal{M}_t = \{i \in \{0,\ldots,N-1\} \mid y_t^i = [\mathrm{MASK}]\}$.
For brevity, we write $p_\theta^i(\cdot \mid \mathbf{I}, \mathbf{x}, \mathbf{y}_t)$ as $p_\theta^i(\cdot \mid \mathbf{y}_t)$.
From this distribution, we obtain $\hat{y}_t^i$ as the token with the highest probability and the confidence $c_t^i$ as its corresponding probability:
\begin{equation}
\hat{y}_t^i
=
\arg\max_{v\in\mathcal{V}}
p_\theta^i(v\mid \mathbf{y}_t),
\qquad
c_t^i
=
p_\theta^i(\hat{y}_t^i\mid \mathbf{y}_t),
\end{equation}
where $\mathcal{V}$ denotes the vocabulary set.
Based on these predictions, a subset $\mathcal{S}\subseteq\mathcal{M}_t$ of $k$ positions is selected, and the selected positions are updated as $y_{t+1}^i = \hat{y}_t^i$ for $i\in\mathcal{S}$.
A common strategy for selecting $\mathcal{S}$ is to choose the masked positions with the highest confidence scores~\cite{llada}.

\paragraph{Attention mechanism in dMLLM.}

At each decoding step $t$, the token embeddings of $\mathbf{I}$, $\mathbf{x}$, and $\mathbf{y}_t$ are concatenated into a single input sequence of length $L$. 
Transformer-based dMLLMs~\cite{li2026lavida, yang2026mmada, llada-v} process this sequence through multiple self-attention layers, enabling tokens to exchange information across modalities.
For each layer and attention head, the self-attention matrix $A\in\mathbb{R}^{L\times L}$ is computed as
\begin{equation}
A
=
\operatorname{softmax}
\left(
QK^\top
\right),
\end{equation}
where $Q$ and $K$ denote the query and key matrices, respectively.
Each element $A_{i,j}$ represents the attention weight from query position $i$ to key position $j$, where a larger value indicates that token $i$ receives more information from token $j$~\cite{vaswani2017attention,zhang2025cross-modal-info-flow}.
Throughout this work, we use $A$ to denote the self-attention matrix from the last layer, averaged over all attention heads.

\paragraph{Information gain in token selection.}
In dLLM decoding, selecting informative tokens for subsequent predictions facilitates the decoding process~\cite{from-bits-to-rounds,info-gain,pc-sampler}.
Such informativeness can be quantified using \textit{information gain}, which measures the reduction in uncertainty of subsequent predictions~\cite{info-gain}.
To formalize this, for a set $\mathcal{S}\subseteq\mathcal{M}_t$, let $\mathcal{M}_t^{(\mathcal{S})}=\mathcal{M}_t\setminus\mathcal{S}$, and let $\mathbf{y}_t^{(\mathcal{S})}$ denote the response obtained by replacing $y_t^i$ with $\hat{y}_t^i$ for $i\in\mathcal{S}$.
For each $\mathbf{y}\in\{\mathbf{y}_t, \mathbf{y}_{t}^{(\mathcal{S})}\}$, we measure this uncertainty as the average entropy over $\mathcal{M}_t^{(\mathcal{S})}$:
\begin{equation}
U\bigl(\mathcal{M}_t^{(\mathcal{S})};\mathbf{y}\bigr)
=
\frac{1}{|\mathcal{M}_t^{(\mathcal{S})}|}
\sum_{i\in\mathcal{M}_t^{(\mathcal{S})}}
H\bigl(p_\theta^i(\cdot\mid\mathbf{y})\bigr),
\end{equation}
where $H(\cdot)$ denotes the entropy of a predicted distribution over tokens.
The information gain $\Delta U_t(\mathcal{S})$ is defined as the reduction in this average entropy after unmasking $\mathcal{S}$: 
\begin{equation}
\Delta U_t(\mathcal{S})
=
U(\mathcal{M}_{t}^{(\mathcal{S})};\mathbf{y}_t)
-
U\bigl(\mathcal{M}_{t}^{(\mathcal{S})};\mathbf{y}_{t}^{(\mathcal{S})}\bigr).
\label{eq}
\end{equation}

\subsection{Motivating Observations}
\label{subsec:motivation}
When generating a response conditioned on multimodal inputs, existing approaches for selecting informative tokens face two practical limitations.
First, evaluating information gain across many candidate token subsets requires an additional forward pass for each subset.
This becomes computationally expensive with long multimodal inputs, as the cost of self-attention grows quadratically with sequence length.
Second, information gain does not necessarily reflect how well a token is grounded in the visual input.
A token can yield substantial information gain solely from linguistic priors or textual dependencies within the response.
To address these limitations, we use the attention weights from masked tokens to image tokens (i.e., image attention) as a readily available signal for assessing token informativeness while accounting for the visual input~\cite{your-vlm-only-needs-few,nguyen2026Beyond-the-Global-Scores,jian2025Look-again-think-slowly}.
We conduct analyses to examine how image attention can guide token selection during dMLLM decoding, leading to the following two observations.

\paragraph{Observation 1: total image attention reflects both visual grounding and token informativeness.}

We examine how the amount of image attention relates to visual grounding and information gain. 
To quantify this amount, we define the image-attention mass as the sum of attention weights over all image tokens.
Throughout the decoding process, we measure the image-attention mass and information gain $\Delta U_t(\{i\})$ for each selected position $i\in\mathcal{S}$.
At the same time, we identify tokens that rely on visual information by examining whether $\hat{y}_t^i$ changes when the image is removed from the input.
A change indicates dependence on visual evidence, in which case we classify the token as \textit{grounded}; otherwise, we classify it as \textit{non-grounded}.
As shown in Figure~\ref{fig:motivation} (top), higher image-attention mass corresponds to both higher information gain and a greater proportion of \textit{grounded} tokens.
This suggests that image-attention mass can serve as a signal for identifying tokens that are both visually grounded and informative.

\paragraph{Observation 2: similar image-attention patterns reflect shared information across tokens.}

In addition to the total amount of image attention, we consider the distribution of attention across image tokens.
Intuitively, tokens with similar image-attention patterns may rely on shared visual evidence and thus provide overlapping information when decoded together.
To quantify this overlap, we define the shared information ratio (SIR) as a measure that compares the information gain from decoding the tokens in $\mathcal{S}$ together with the sum of the information gains obtained by decoding each token individually:
\begin{equation}
    \mathrm{SIR}_t(\mathcal{S})
    =
    1
    -
    \frac{
    \Delta U_t(\mathcal{S})
    }{
    \sum_{i\in\mathcal{S}}\Delta U_t(\{i\})
    }.
    \label{eq:wasted_information_ratio}
\end{equation}
A larger $\mathrm{SIR}_t(\mathcal{S})$ indicates a greater degree of information overlap among the tokens in $\mathcal{S}$.
We then examine the relationship between image-attention similarity and SIR.
For each subset size $k\in\{2,4,8\}$, we decode $k$ tokens per step and compute the SIR and mean pairwise image-attention similarity of the selected subset $\mathcal{S}$.
We then report the average SIR within each image-attention similarity range.
As shown in Figure~\ref{fig:motivation} (bottom), SIR increases with image-attention similarity, consistent with the intuition that similar attention patterns reflect shared information across tokens.
This suggests that image-attention distributions can provide a cue for selecting token sets that yield higher information gain when decoded together.
Additional experimental details for these observations are provided in the Appendix.

\begin{table*}[t]
\centering
\setlength{\tabcolsep}{1.0pt}
\renewcommand{\arraystretch}{1.0}

\begin{tabular*}{\textwidth}{
@{\extracolsep{\fill}}
c l
c c c c
c
c c c c
@{}
}
\toprule
$k$
& \textbf{Sampler}
& \textbf{COCO Cap.}
& \textbf{Flickr30K}
& \textbf{NoCaps}
& \textbf{Avg.}
& \textbf{DetailCaps}
& \textbf{TextVQA}
& \textbf{DocVQA}
& \textbf{ChartQA}
& \textbf{Avg.}
\\[-1pt]

\multicolumn{2}{c}{}
& {\scriptsize\textit{CIDEr}}
& {\scriptsize\textit{CIDEr}}
& {\scriptsize\textit{CIDEr}}
& {\scriptsize\textit{CIDEr}}
& {\scriptsize\textit{CAPTURE}}
& {\scriptsize\textit{Acc.}}
& {\scriptsize\textit{ANLS}}
& {\scriptsize\textit{Relaxed Acc.}}
& {\scriptsize\textit{Macro Avg.}}
\\
\midrule

\multirow{7}{*}{2}
& Confidence    & 100.1 & 69.7 & 88.4 & 86.1 & 56.2 & 55.2 & 60.1 & 56.0 & 57.1 \\
& Entropy       & 95.4  & 68.2 & 83.9 & 82.5 & 56.5 & 54.6 & 59.8 & 56.0 & 56.8 \\
& Margin        & 105.8 & \underline{71.7} &88.7 & 88.7 & 56.9 & 55.0 & 61.6 & \underline{56.6} & 57.7 \\
& MPD-PAC           & 95.1  & 70.1 & 87.2 & 84.1 & 56.9 & 56.0 & \textbf{62.8} & 56.2 & 58.3 \\
& PC-Sampler    & 101.7 & 69.8 & 88.9 & 86.8 & \underline{57.1} & 54.3 & 58.9 & 55.0 & 56.1 \\
& Info-Gain     & \underline{106.0} & 71.2 & \underline{89.6} & \underline{88.9} & 55.6 & \underline{57.4} & 61.4 & \textbf{57.0} & \underline{58.6} \\
\cmidrule(lr){2-11}
& \textbf{VIG-Sampler}  & \textbf{106.2} & \textbf{74.7} & \textbf{90.5} & \textbf{90.5} & \textbf{57.7} & \textbf{58.7} & \underline{62.6} & 55.8 & \textbf{59.0} \\
\midrule

\multirow{7}{*}{4}
& Confidence    & 93.2 & 63.8 & 78.2 & 78.4 & 49.5 & 47.4 & 59.6 & 53.0 & 53.3 \\
& Entropy       & 91.2 & 63.5 & 78.4 & 77.7 & 45.9 & 46.0 & 57.7 & 52.6 & 52.1 \\
& Margin        & 93.7 & 65.5 & 78.0 & 79.1 & \underline{51.5} & 47.4 & 59.7 & 53.2 & 53.4 \\
& MPD-PAC           & 92.4  & 65.5 & 79.4 & 79.1 & 49.0 & 47.9 & \underline{60.5} & 53.6 & 54.0 \\
& PC-Sampler    & \underline{95.8} & \underline{66.6} & \underline{82.8} & \underline{81.7} & 49.6 & \underline{52.2} & 57.1 & \underline{54.0} & 54.4 \\
& Info-Gain     & 94.8 & 65.9 & 81.9 & 80.9 & 49.2 & 50.2 & 60.2 & \underline{54.0} & \underline{54.8} \\
\cmidrule(lr){2-11}
& \textbf{VIG-Sampler}  & \textbf{105.1} & \textbf{71.2} & \textbf{87.4} & \textbf{87.9} & \textbf{55.9} & \textbf{56.8} & \textbf{62.3} & \textbf{55.8} & \textbf{58.3} \\
\midrule

\multirow{7}{*}{8}
& Confidence    & 74.5 & 49.0 & 58.9 & 60.8 & 29.2 & 41.7 & 54.1 & 50.0 & 48.6 \\
& Entropy       & 71.6 & 49.4 & 58.7 & 59.9 & 27.4 & 41.7 & 53.8 & 49.8 & 48.4 \\
& Margin        & 73.2 & 46.9 & 59.8 & 60.0 & \underline{34.0} & 41.7 & 54.4 & 50.0 & 48.7 \\
& MPD-PAC           & 73.4  & 49.0 & 59.7 & 60.7 & 29.9 & 40.7 & 55.7 & 50.4 & 48.9 \\
& PC-Sampler    & \underline{83.9} & \underline{56.2} & \underline{67.9} & \underline{69.3} & 30.2 & \underline{50.8} & \underline{56.8} & \underline{54.0} & \underline{53.9} \\
& Info-Gain     & 76.7 & 50.6 & 60.9 & 62.7 & 29.5 & 42.7 & 54.4 & 49.8 & 49.0 \\
\cmidrule(lr){2-11}
& \textbf{VIG-Sampler}  & \textbf{100.1} & \textbf{63.6} & \textbf{82.3} & \textbf{82.0} & \textbf{42.4} & \textbf{53.1} & \textbf{61.5} & \textbf{54.2} & \textbf{56.3} \\

\bottomrule
\end{tabular*}

\caption{
Experimental results of LaViDa with parallel decoding budgets $k \in \{2,4,8\}$.
The first and last Avg.\ columns report mean CIDEr and VQA scores, respectively.
The best scores are shown in \textbf{bold}, and the second-best scores are \underline{underlined}.
}
\label{tab:main_results_1}
\vspace{-6pt}
\end{table*}

\subsection{VIG-Sampler}

Motivated by these observations, we propose VIG-Sampler, a visually grounded decoding strategy for dMLLMs that leverages image attention as an additional cue.
Specifically, VIG-Sampler guides token selection by prioritizing visually grounded and informative tokens based on image-attention mass, while penalizing high image-attention similarity within the selected set to reduce information overlap.
An overview of VIG-Sampler is provided in Figure~\ref{fig:method}, with the overall decoding procedure detailed in the Appendix.

\paragraph{Score reweighting with image-attention mass.}
Based on Observation~1, we use image-attention mass to reweight the confidence score $c_t^i$ used for masked-token selection.
For each masked position $i\in\mathcal{M}_t$, let $\mathbf{a}_t^i$ denote the attention weights from position $i$ to the image-token positions $\mathcal{I}$, and let $m_t^i$ denote the image-attention mass:
\begin{equation}
    \mathbf{a}_t^i
    =
    A_{i,\mathcal{I}},
    \qquad
    m_t^i
    =
    \lVert \mathbf{a}_t^i \rVert_1
    =
    \sum_{j\in\mathcal{I}} A_{i,j}.
    \label{eq:image_attention}
\end{equation}
To reflect the relative magnitude of $m_t^i$, we normalize it by $m_t^{\mathrm{med}}$, the median image-attention mass over all masked positions.
We then use the normalized image-attention mass to reweight the confidence score:
\begin{equation}
    r_t^i = c_t^i \left( \frac{m_t^i}{m_t^{\mathrm{med}}} \right)^{\gamma},
    \label{eq:reward_guided_score}
\end{equation}
where $\gamma\geq 0$ is a hyperparameter that controls the strength of image-attention guidance.
This formulation gradually incorporates image-attention guidance into the confidence score as $\gamma$ increases, starting from $r_t^i=c_t^i$ at $\gamma=0$.
For $\gamma>0$, positions with higher image-attention mass receive larger weights, thereby prioritizing tokens that attend more strongly to the image.
In this way, VIG-Sampler favors visually grounded and informative tokens while retaining the model’s original preference (i.e., confidence-based preference). 
The resulting scores are then used in the subsequent set-selection objective.

\begin{table*}[t]
\centering
\setlength{\tabcolsep}{1.0pt}
\renewcommand{\arraystretch}{1.0}

\begin{tabular*}{\textwidth}{
@{\extracolsep{\fill}}
c l
c c c c
c
c c c c
@{}
}
\toprule
$k$
& \textbf{Sampler}
& \textbf{COCO Cap.}
& \textbf{Flickr30K}
& \textbf{NoCaps}
& \textbf{Avg.}
& \textbf{DetailCaps}
& \textbf{TextVQA}
& \textbf{DocVQA}
& \textbf{ChartQA}
& \textbf{Avg.}
\\[-1pt]

\multicolumn{2}{c}{}
& {\scriptsize\textit{CIDEr}}
& {\scriptsize\textit{CIDEr}}
& {\scriptsize\textit{CIDEr}}
& {\scriptsize\textit{CIDEr}}
& {\scriptsize\textit{CAPTURE}}
& {\scriptsize\textit{Acc.}}
& {\scriptsize\textit{ANLS}}
& {\scriptsize\textit{Relaxed Acc.}}
& {\scriptsize\textit{Macro Avg.}}
\\
\midrule

\multirow{7}{*}{2}
& Confidence    & 87.4 & 57.7 & 86.5 & 77.2 & 50.1 & 12.2 & 9.3 & \underline{11.8} & 11.1 \\
& Entropy       & 82.5 & 55.8 & 83.3 & 73.9 & \textbf{50.5} & 12.0 & 8.8 & \underline{11.8} & 10.9 \\
& Margin        & \underline{88.0} & 58.7 & \underline{88.0} & 78.2 & \textbf{50.5} & 12.2 & 9.5 & 11.6 & 11.1 \\
& MPD-PAC           & 86.1  & 56.7 & 81.2 & 74.7 & 50.0 & \underline{12.4} & 10.1 & 11.6 & 11.4 \\
& PC-Sampler    & 87.1 & 57.4 & 80.3 & 74.9 & 49.3 & \underline{12.4} & 8.6 & \underline{11.8} & 10.9 \\
& Info-Gain & \textbf{89.1} & \underline{59.8} & 86.9 & \underline{78.6} & 48.8 & \textbf{12.8} & \textbf{10.6} & \underline{11.8} & \textbf{11.7} \\
\cmidrule(lr){2-11}
& \textbf{VIG-Sampler} & 87.9 & \textbf{60.2} & \textbf{89.2} & \textbf{79.1} & \underline{50.4} & \underline{12.4} & \underline{10.5} & \textbf{12.0} & \underline{11.6} \\
\midrule

\multirow{7}{*}{4}
& Confidence    & 83.4 & 56.8 & 78.2 & 72.8 & 45.5 & 11.1 & 7.9 & 12.0 & 10.3 \\
& Entropy       & 78.8 & 54.7 & 75.9 & 69.8 & 44.5 & 11.1 & 7.6 & 12.0 & 10.2 \\
& Margin        & 81.1 & 57.2 & 80.7 & 73.0 & \underline{46.6} & 11.1 & 8.0 & 11.6 & 10.2 \\
& MPD-PAC           & 80.9 & 52.4 & 77.3 & 70.2 & 46.5 & 11.4 & 8.3 & \underline{12.2} & 10.6 \\
& PC-Sampler    & 81.5 & 51.4 & 73.5 & 68.8 & 45.8 & \underline{12.4} & 8.6 & 12.0 & \underline{11.0} \\
& Info-Gain     & \underline{83.8} & \underline{58.0} & \underline{81.4} & \underline{74.4} & 45.8 & 11.4 & \underline{8.9} & 12.0 & 10.8 \\
\cmidrule(lr){2-11}
& \textbf{VIG-Sampler}  & \textbf{87.6} & \textbf{58.8} & \textbf{86.6} & \textbf{77.7} & \textbf{49.1} & \textbf{12.7} & \textbf{9.9} & \textbf{13.2} & \textbf{11.9} \\
\midrule

\multirow{7}{*}{8}
& Confidence    & 61.6 & 41.7 & 61.6 & 55.0 & 33.1 & 11.0 & 8.4 & \underline{12.2} & 10.5 \\
& Entropy       & 60.7 & 41.0 & 60.1 & 53.9 & 31.8 & 11.0 & 8.3 & \underline{12.2} & 10.5 \\
& Margin        & \underline{66.4} & \underline{44.8} & 64.0 & \underline{58.4} & 35.7 & 11.0 & 8.5 & \underline{12.2} & 10.6 \\
& MPD-PAC           & 61.0  & 41.2 & 60.2 & 54.1 & \underline{38.6} & 11.1 & \underline{8.6} & \underline{12.2} & 10.6 \\
& PC-Sampler    & 62.9 & 43.1 & 62.8 & 56.3 & 33.5 & \textbf{12.6} & 8.4 & 12.0 & \underline{11.0} \\
& Info-Gain     & 65.4 & 43.8 & \underline{65.2} & 58.1 & 33.1 & 11.0 & 8.4 & 11.8 & 10.4 \\
\cmidrule(lr){2-11}
& \textbf{VIG-Sampler}  & \textbf{81.9} & \textbf{58.3} & \textbf{78.4} & \textbf{72.9} & \textbf{43.3} & \underline{12.3} & \textbf{9.5} & \textbf{12.4} & \textbf{11.4} \\
\bottomrule

\end{tabular*}
\caption{
Experimental results of MMaDA with parallel decoding budgets $k \in \{2,4,8\}$.
The first and last Avg.\ columns report mean CIDEr and VQA scores, respectively.
The best scores are shown in \textbf{bold}, and the second-best scores are \underline{underlined}.
}
\label{tab:main_results_2}
\vspace{-6pt}
\end{table*}

\paragraph{Set selection via image-attention similarity regularization.}
\label{subsec:set_selection}

Building on Observation~2, we use pairwise image-attention similarity as a regularization term when selecting the token set $\mathcal{S}$ at each decoding step.
To better capture how each image-attention pattern differs from the overall pattern, we center them by subtracting the mean over $\mathcal{M}_t$:
\begin{equation}
\tilde{\mathbf{a}}_t^i
=
\mathbf{a}_t^i
-
\frac{1}{|\mathcal{M}_t|}
\sum_{j\in\mathcal{M}_t}
\mathbf{a}_t^j.
\label{eq:centered_image_attention}
\end{equation}
Using $\tilde{\mathbf{a}}_t^i$, we compute the pairwise image-attention similarity as
\begin{equation}
G_t^{i,j}
=
\max
\left\{
\left\langle
\tilde{\mathbf{a}}_t^i,
\tilde{\mathbf{a}}_t^j
\right\rangle_{\mathrm{cos}},
0
\right\},
\label{eq:image_attention_similarity}
\end{equation}
where $\langle\cdot,\cdot\rangle_{\mathrm{cos}}$ denotes cosine similarity.
We then formulate the set-selection objective based on the reweighted token scores while accounting for pairwise image-attention similarity:
\begin{equation}
\mathcal{S}^\star_t
=
\arg\max_{\substack{
\mathcal{S}\subseteq\mathcal{M}_t\\
|\mathcal{S}|=k
}}
\left[
\sum_{i\in\mathcal{S}}r_t^i
-
\frac{\lambda}{|\mathcal{S}|-1}
\sum_{\substack{i,j\in\mathcal{S}\\ i<j}}
G_t^{i,j}
\right],
\label{eq:set_selection}
\end{equation}
where $\lambda\geq0$ is a hyperparameter controlling the strength of the similarity regularization.
The second term is defined as zero for $|\mathcal{S}|=1$.
Since $G_t^{i,j}$ is non-negative, it acts solely as a penalty for similar image-attention patterns rather than rewarding dissimilar ones.
VIG-Sampler thus discourages selecting tokens with similar image-attention patterns, reducing information overlap within the selected set.

Finding the optimal set $\mathcal{S}_t^\star$ in Eq.~\eqref{eq:set_selection} requires searching over a combinatorial number of candidate subsets, which is computationally impractical.
We therefore use a greedy algorithm that selects the next position to maximize the marginal gain in the objective given the currently selected set (see the Appendix for its derivation).
After initializing $\mathcal{S}$ to $\emptyset$, we select the first position solely based on $r_t^i$.
We then select the next position $i^\star$ as
\begin{equation}
i^\star
=
\arg\max_{i\in\mathcal{M}_t^{(\mathcal{S})}}
\left[
r_t^i
-
\frac{\lambda}{|\mathcal{S}|}
\sum_{j\in\mathcal{S}}
G_t^{i,j}
\right].
\label{eq:greedy_selection}
\end{equation}
We iteratively add the selected position $i^\star$ to $\mathcal{S}$ until $|\mathcal{S}|=k$.
The resulting set $\mathcal{S}$ determines the positions to be decoded with their predicted tokens $\hat{y}_t^i$ at the current step.

\section{Experiments}

\subsection{Models and Baselines}

Our evaluation covers three representative dMLLMs: LaViDa-LLaDA-v1.0-Instruct~\cite{li2026lavida}, MMaDA-8B-MixCoT~\cite{yang2026mmada}, and LLaDA-V~\cite{llada-v}.
Across these models, we evaluate VIG-Sampler against six sampling baselines: Confidence~\cite{chang2022maskgit}, Entropy~\cite{ebsampler}, Margin~\cite{kim2025train}, MPD-PAC~\cite{hong2026mitigating}, PC-Sampler~\cite{pc-sampler}, and Info-Gain~\cite{info-gain}.
The first three are widely used uncertainty-based heuristics, whereas MPD-PAC is designed to enhance visual grounding in dMLLMs.
PC-Sampler incorporates global trajectory and content-aware guidance, whereas Info-Gain accounts for the future uncertainty reduction induced by each decoding decision.
Note that we use fixed hyperparameters of $\gamma=1$ and $\lambda=3$ for all models.
Further implementation and benchmark details are provided in the Appendix.

\begin{figure*}[t!]
  \centering
  \includegraphics[width=\textwidth]{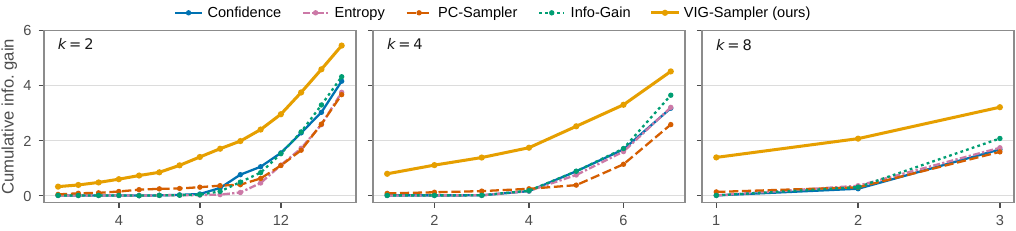}
  \vspace{-2pt}
    \caption{
Cumulative information gain across decoding steps for $k\in\{2,4,8\}$
    }
  \label{fig:information_gain}
\end{figure*}

\begin{figure*}[t!]
  \centering
  \includegraphics[width=\textwidth]{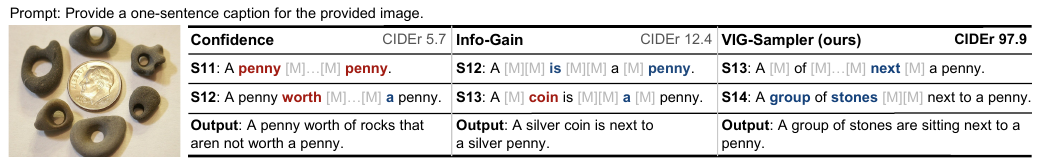}
  \vspace{-1pt}
    \caption{Qualitative comparison on COCO Caption using LaViDa at $k=2$.
    \textbf{S\#} denotes the decoding step, and red and blue indicate incorrectly and correctly committed tokens, respectively.}
  \label{fig:qualitative_1}
\end{figure*}

\subsection{Main Results}

Tables~\ref{tab:main_results_1} and~\ref{tab:main_results_2} summarize the experimental results on LaViDa and MMaDA, respectively.
VIG-Sampler achieves the highest mean CIDEr score across all model--budget settings and obtains the best DetailCaps and VQA averages in five of the six settings, with the second-best result in the remaining setting.
Its consistent gains across captioning and VQA suggest that visual reward-guided token scoring effectively prioritizes visually grounded and informative tokens.
The advantage of VIG-Sampler becomes more pronounced as the decoding budget increases.
At $k=8$, it outperforms the second-best baseline by $12.7$ and $14.5$ in mean CIDEr score on LaViDa and MMaDA, respectively, with corresponding CAPTURE gains of $8.4$ and $4.7$.
Remarkably, VIG-Sampler at $k=8$ matches or outperforms certainty-based sampling at $k=2$ on COCO Caption with LaViDa and Flickr30K with MMaDA, despite a $4\times$ reduction in decoding steps.
Together, these results show that joint-information-aware set selection enables VIG-Sampler to gain more information at each step, benefiting from larger commit budgets while preserving generation quality.

\section{Analysis}

\subsection{Analysis of VIG-Sampler}

\paragraph{Information gain induced by VIG-Sampler.}

Figure~\ref{fig:information_gain} compares the cumulative information gain of VIG-Sampler against competing sampling baselines across decoding steps on COCO Caption using LaViDa.
To better isolate information gain related to visual content, we exclude punctuation, articles, and other function words, whose predictions are more strongly driven by linguistic context than by visual evidence.
Although VIG-Sampler does not directly compute information gain, it consistently achieves the highest cumulative information gain across all commit budgets.

\paragraph{Qualitative case studies.}
Figure~\ref{fig:qualitative_1} shows that Confidence and Info-Gain commit tokens that convey overlapping information, such as \textit{``penny''} and \textit{``coin''}, without accounting for the joint information gain of the selected set.
These early commitments bias subsequent predictions toward plausible but incorrect continuations, yielding phrases such as \textit{``penny worth''} and semantically repetitive descriptions such as \textit{``silver coin''} and \textit{``silver penny''}.
In contrast, VIG-Sampler selects complementary, visually grounded tokens and generates a caption consistent with the image.

\subsection{Ablation Study}

\paragraph{Hyperparameter sensitivity.}
Table~\ref{tab:gamma_lambda_ablation} presents a sensitivity analysis of $\gamma$ and $\lambda$ on COCO Caption with LaViDa at $k=2$.
Performance generally improves as either $\gamma$ or $\lambda$ increases from zero, and their gains are largely additive when the two are combined.
Performance remains robust across a broad range of values; although some configurations outperform the default, we fix $\gamma=1$ and $\lambda=3$ across all models and benchmarks for consistency.

\paragraph{Robustness to generation length.}

Table~\ref{tab:gen_length_ablation_k4} compares VIG-Sampler and Info-Gain across different generation lengths at $k=4$.
VIG-Sampler consistently outperforms Info-Gain on both COCO Caption and TextVQA for all values of $N$, demonstrating robustness across generation lengths.

\begin{table}[t]
\centering
\setlength{\tabcolsep}{6pt}
\renewcommand{\arraystretch}{1.0}

\begin{tabular}{c|ccccc}
\toprule
$\gamma \backslash \lambda$
& $0.0$ & $1.0$ & $2.0$ & $3.0$ & $4.0$ \\
\midrule
$0.0$ & 100.2 & 100.9 & 102.1 & 102.9 & 101.7 \\
$0.5$ & 101.9 & 105.2 & 106.9 & 106.5 & 105.2 \\
$1.0$ & 101.5 & 103.8 & 105.4 & 106.2$^{\dagger}$ & 105.5 \\
$1.5$ & 101.0 & 103.5 & 106.0 & 107.7 & \textbf{107.8} \\
$2.0$ & 103.0 & 107.0 & 106.2 & 106.1 & 107.5 \\
\bottomrule
\end{tabular}
\caption{
Hyperparameter sensitivity on COCO Caption using LaViDa.
$^{\dagger}$ denotes the default setting.
}
\label{tab:gamma_lambda_ablation}
\end{table}

\begin{table}[t]
\centering
\renewcommand{\arraystretch}{0.8}
\begin{tabular}{c|c|cc}
\toprule
$N$ & Sampler & COCO Cap. & TextVQA \\
\midrule

\multirow{2}{*}{16}
& Info-Gain    & 81.6  & 48.8 \\
& VIG-Sampler  & \textbf{88.4}  & \textbf{55.0} \\
\cmidrule{1-4}

\multirow{2}{*}{32}
& Info-Gain    & 94.8  & 50.2 \\
& VIG-Sampler  & \textbf{105.1} & \textbf{56.8} \\
\cmidrule{1-4}

\multirow{2}{*}{48}
& Info-Gain    & 97.4  & 48.6 \\
& VIG-Sampler  & \textbf{102.5} & \textbf{57.4} \\
\cmidrule{1-4}

\multirow{2}{*}{64}
& Info-Gain    & 96.3  & 49.5 \\
& VIG-Sampler  & \textbf{103.7} & \textbf{55.8} \\
\bottomrule
\end{tabular}
\caption{Robustness to $N$ using LaViDa at $k=4$.}
\label{tab:gen_length_ablation_k4}
\end{table}

\section{Conclusion}
In this paper, we presented VIG-Sampler, a training-free parallel decoding strategy for dMLLMs that leverages token-to-image attention to guide token selection.
By combining visual reward-guided token scoring with joint-information-aware set selection, VIG-Sampler prioritizes visually grounded and informative tokens during parallel decoding.
Experiments across seven captioning and VQA benchmarks and three dMLLMs demonstrate consistent improvements over existing sampling baselines, with particularly strong gains under larger commit budgets.
These results highlight the importance of incorporating visual information into the decoding order to preserve generation quality with fewer decoding steps.

\clearpage

\bibliography{aaai2027}


\newpage
\appendix
\section{Related Work}

\subsection{Diffusion Multimodal Large Language Models}
Multimodal large language models (MLLMs) project visual information into the representation space of a pretrained LLM, achieving strong performance on multimodal tasks such as captioning, visual question answering, and visual reasoning~\cite{liu2023visual, liu2024improved, dai2023instructblip}.
With the emergence of dLLMs as an alternative to autoregressive LLMs~\cite{llada, gong2025scaling, ye2025dream}, this paradigm has also been extended to multimodal models that denoise sequences conditioned on both visual and textual tokens~\cite{llada-v, li2026lavida, yu2025dimple}.
More recent models jointly learn visual and textual tokens in a unified framework, rather than adapting visual features to language space, enabling balanced multimodal understanding and visual generation~\cite{yang2026mmada, li2025lavida-o, shi2025muddit}.

Recent studies on dMLLM decoding have focused on improving inference efficiency by pruning redundant masked or visual tokens during generation~\cite{li2025comprehensive, li2025sparse, xu2025redvtp}.
Another line of work improves the visual grounding of generated responses by adjusting token-level predictions through model-internal bias correction or classifier-free guidance~\cite{hong2026mitigating, kim2026thinking}. 
Despite the growing body of research on dMLLMs, relatively little attention has been paid to which tokens should be decoded at each step and how these decoding decisions affect the subsequent generation trajectory.

\subsection{Parallel Decoding Strategies in Diffusion Language Models}
At each denoising step, diffusion language models predict all remaining masked positions and commit a subset of them for parallel decoding.
The most common strategies select positions based on predictive certainty, measured by confidence~\cite{llada, li2026lavida}, entropy~\cite{ye2025dream, ebsampler}, or the probability margin between the top two candidates~\cite{kim2025train, li2025diffusion}.
Based on these signals, several methods dynamically determine which and how many tokens to commit using threshold-based criteria~\cite{wu2025fast, yu2025dimple, ebsampler}.
Other training-free approaches leverage signals accumulated across denoising steps, such as changes in predictive distributions, convergence, and prediction stability~\cite{kim2026klass, wei2025accelerating}.
Another class of approaches learns token selection policies from trace-level signals, such as future-stability labels derived from completed decoding traces or rewards based on final outputs~\cite{bao2025learning, jazbec2025learning}.
Recent studies have also explored alternative criteria, including information gain over the remaining masked positions, semantic priors, and search over candidate decoding trajectories~\cite{info-gain,pc-sampler,lee2025lookUM}.
Although these methods are generally effective in language-only settings, parallel decoding strategies for multimodal language models remain relatively underexplored, especially those that use visual grounding to guide token selection.

\section{Additional Details}

\subsection{Experimental Details for the Motivation Study}

We conduct all experiments in the motivation study on a subset of 100 images from the COCO Captions validation set.
For observation~1, we first generate captions using the confidence-based sampler with $k=1$.
For each generated caption, we mask one token at a time and run the model again with the image embeddings replaced by zero embeddings.
If the prediction at the masked position changes, we classify the token as \textit{visually grounded}; otherwise, we classify it as \textit{non-grounded}.
The image-attention mass and image-attention similarity used in the experiments are defined as $m_t^i$ and $\langle\tilde{\mathbf{a}}_t^i,\tilde{\mathbf{a}}_t^j\rangle_{\mathrm{cos}}$, respectively.

\subsection{Benchmarks}

We evaluate VIG-Sampler on four image captioning benchmarks and three visual question answering benchmarks, covering a broad range of vision-language capabilities.
For image captioning, we use COCO Caption~\cite{lin2014microsoft}, Flickr30K~\cite{young2014image}, NoCaps~\cite{agrawal2019nocaps}, and DetailCaps~\cite{dong2024benchmarking}, which assess general image description, open-vocabulary generalization, and fine-grained, long-form visual description.
For visual question answering, we use TextVQA~\cite{singh2019towards}, DocVQA~\cite{mathew2020docvqa}, and ChartQA~\cite{masry2022chartqa}, covering scene text understanding, document comprehension, and chart reasoning, respectively.
We report CIDEr for COCO Caption, Flickr30K, and NoCaps, CAPTURE for DetailCaps, and accuracy, ANLS, and relaxed accuracy for TextVQA, DocVQA, and ChartQA, respectively.
We additionally report the mean CIDEr score across the first three captioning benchmarks and the mean score across the three VQA benchmarks.
For a consistent evaluation protocol, we use the \texttt{lmms-eval} package~\cite{zhang2025lmms}, evaluating DetailCaps on 500 samples and the remaining benchmarks on their lite subsets.

\subsection{Implementation Details}
\paragraph{Sampler configurations.}
We retain the default hyperparameter settings reported for prior methods whenever possible and make only minimal adjustments when differences in generation length or dataset characteristics lead to unstable decoding behavior.
For MPD-PAC, we set the RoPE coefficient to $\beta=0.01$, the threshold to $\tau_0=0.6$, $k=3$, and the slope of the sigmoid schedule to $8$.
We use $\lambda_{\mathrm{prior}}=0.3$ for LaViDa, $\lambda_{\mathrm{prior}}=0.1$ for LLaDA-V, and $\lambda_{\mathrm{prior}}=0.03$ for MMaDA.
For PC-Sampler, we set the positional decay coefficient to $\lambda=0.025$ and set the clipping threshold to $\alpha=10$.
For Info-Gain, we use a position temperature of $\tau_{\mathrm{pos}}=0.1$, $N=4$ candidates, and an acceleration threshold of $0.8$.
For VIG-Sampler, we set $\gamma=1$ and $\lambda=3$.

\paragraph{Generation settings.}
For LaViDa and MMaDA, we use a generation length of $N=32$ on all benchmarks except DetailCaps.
On DetailCaps, we employ block decoding for both models with a generation length of $N=128$ and a block length of $16$.
For LLaDA-V, we use $N=16$ on COCO Caption, Flickr30K, and NoCaps, $N=128$ on DetailCaps, and $N=32$ on the VQA benchmarks.

\paragraph{Evaluation setup.}
All experiments are conducted on NVIDIA RTX A5000 GPUs.
We set the sampling temperature to zero and report single-run results.

\subsection{Derivation of the Greedy Selection Rule}

We derive the greedy rule used to approximately optimize the
set-level objective in Eq.~\eqref{eq:set_selection}.
For the current set $\mathcal{S}$ of $n$ selected positions, i.e., $|\mathcal{S}|=n$, define
\begin{equation}
F_t(\mathcal{S})
=
\sum_{i\in\mathcal{S}} r_t^i
-
\frac{\lambda}{|\mathcal{S}|-1}
\sum_{\substack{i,j\in\mathcal{S}\\ i<j}}
G_t^{i,j},
\label{eq:appendix_set_objective}
\end{equation}
where the pairwise similarity term is defined as zero when
$|\mathcal{S}|=1$, and we set $F_t(\emptyset)=0$.
Consider adding a candidate
$i\in\mathcal{M}_t^{(\mathcal{S})}$.
Define the accumulated pairwise similarity within the current set as
\begin{equation}
P_t(\mathcal{S})
=
\sum_{\substack{a,b\in\mathcal{S}\\ a<b}}
G_t^{a,b}.
\label{eq:current_pairwise_similarity}
\end{equation}
After adding candidate $i$, the pairwise similarity term additionally
contains its similarities to all positions already in $\mathcal{S}$.
Thus, for $n\geq 2$,
\begin{align}
F_t(\mathcal{S}\cup\{i\})
&=
\sum_{j\in\mathcal{S}} r_t^j
+
r_t^i
\nonumber\\
&\quad
-
\frac{\lambda}{n}
\left[
P_t(\mathcal{S})
+
\sum_{j\in\mathcal{S}}G_t^{i,j}
\right].
\label{eq:objective_after_addition}
\end{align}
Subtracting the current objective
\begin{equation}
F_t(\mathcal{S})
=
\sum_{j\in\mathcal{S}}r_t^j
-
\frac{\lambda}{n-1}P_t(\mathcal{S})
\end{equation}
gives the marginal gain
\begin{align}
F_t(\mathcal{S}\cup\{i\})-F_t(\mathcal{S})
&=
r_t^i
-
\frac{\lambda}{n}
\sum_{j\in\mathcal{S}}G_t^{i,j}
\nonumber\\
&\quad
+
\frac{\lambda}{n(n-1)}
P_t(\mathcal{S}).
\label{eq:marginal_gain}
\end{align}
The last term depends only on the currently selected set
$\mathcal{S}$ and is independent of candidate $i$. Therefore, it does
not affect which candidate maximizes the marginal gain.

For $n=1$, the current set has no pairwise similarity term, and adding
$i$ yields
\begin{equation}
F_t(\mathcal{S}\cup\{i\})-F_t(\mathcal{S})
=
r_t^i
-
\lambda
\sum_{j\in\mathcal{S}}G_t^{i,j},
\end{equation}
which has the same candidate-dependent form as
Eq.~\eqref{eq:marginal_gain} with $n=1$.

Consequently, for any nonempty $\mathcal{S}$, the greedy selection rule
is
\begin{equation}
i^\star
=
\arg\max_{i\in\mathcal{M}_t^{(\mathcal{S})}}
\left[
r_t^i
-
\frac{\lambda}{|\mathcal{S}|}
\sum_{j\in\mathcal{S}}
G_t^{i,j}
\right].
\label{eq:derived_greedy_selection}
\end{equation}
When $\mathcal{S}=\emptyset$, the similarity term is defined as zero,
so the first selected position is
\begin{equation}
i^\star
=
\arg\max_{i\in\mathcal{M}_t} r_t^i.
\end{equation}

\begin{table*}[t]
\centering
\setlength{\tabcolsep}{2.5pt}
\begin{tabular*}{\textwidth}{
@{\extracolsep{\fill}}
c l
c c c c
c
c c c c
@{}
}
\toprule
$k$
& \textbf{Sampler}
& \textbf{COCO Cap.}
& \textbf{Flickr30K}
& \textbf{NoCaps}
& \textbf{Avg.}
& \textbf{DetailCaps}
& \textbf{TextVQA}
& \textbf{DocVQA}
& \textbf{ChartQA}
& \textbf{Avg.}
\\[-1pt]
\multicolumn{2}{c}{}
& {\scriptsize\textit{CIDEr}}
& {\scriptsize\textit{CIDEr}}
& {\scriptsize\textit{CIDEr}}
& {\scriptsize\textit{CIDEr}}
& {\scriptsize\textit{CAPTURE}}
& {\scriptsize\textit{Acc.}}
& {\scriptsize\textit{ANLS}}
& {\scriptsize\textit{Relaxed Acc.}}
& {\scriptsize\textit{Macro Avg.}}
\\
\midrule

\multirow{7}{*}{2}
& Confidence    & 92.8 & \underline{83.9} & 91.6 & 89.4 & 59.9 & \underline{67.3} & 81.1 & 76.2 & 74.9 \\
& Entropy       & 91.0 & 77.8 & 89.4 & 86.1 & 59.0 & 63.9 & 82.0 & 76.6 & 74.2 \\
& Margin        & \textbf{94.5} & 82.5 & \underline{91.8} & \underline{89.6} & \textbf{61.1} & \underline{67.3} & 82.0 & \textbf{77.6} & \underline{75.6} \\
& MPD-PAC       & 92.4 & 83.6 & 91.2 & 89.1 & \underline{60.5} & 65.0 & 81.4 & 75.8 & 74.1 \\
& PC-Sampler    & 92.1 & 81.9 & 90.8 & 88.3 & 59.8 & 66.5 & \underline{82.8} & 76.4 & 75.2 \\
& Info-Gain     & \underline{94.4} & \textbf{86.1} & \textbf{93.8} & \textbf{91.4} & 60.1 & 67.2 & 82.0 & 76.6 & 75.3 \\
\cmidrule(lr){2-11}
& \textbf{VIG-Sampler}  & 93.9 & 82.5 & 91.3 & 89.2 & 59.9 & \textbf{67.8} & \textbf{83.4} & \underline{77.0} & \textbf{76.1} \\
\midrule

\multirow{7}{*}{4}
& Confidence    & 89.4 & 77.9 & 84.9 & 84.1 & 58.4 & 63.2 & 80.2 & 75.6 & 73.0 \\
& Entropy       & 85.7 & 73.3 & 83.8 & 80.9 & 56.3 & 62.2 & 78.6 & 73.4 & 71.4 \\
& Margin        & 88.9 & 76.6 & 85.6 & 83.7 & \underline{59.9} & \textbf{63.8} & \underline{81.4} & \textbf{76.4} & \underline{73.9} \\
& MPD-PAC       & 89.0 & 76.0 & 84.2 & 83.1 & 58.8 & 62.5 & 79.5 & 74.8 & 72.3 \\
& PC-Sampler    & 85.6 & 74.2 & 81.9 & 80.6 & 53.9 & \underline{63.6} & \textbf{82.6} & \underline{76.0} & \textbf{74.1} \\
& Info-Gain     & \textbf{90.8} & \underline{78.1} & \underline{88.0} & \underline{85.6} & 58.6 & 62.1 & 80.7 & 75.8 & 72.9 \\
\cmidrule(lr){2-11}
& \textbf{VIG-Sampler}  & \underline{90.7} & \textbf{78.5} & \textbf{90.4} & \textbf{86.5} & \textbf{60.3} & \textbf{63.8} & 79.6 & \textbf{76.4} & 73.3 \\
\midrule

\multirow{7}{*}{8}
& Confidence    & 72.3 & 61.4 & 66.8 & 66.8 & 49.0 & 51.2 & 73.3 & 72.0 & 65.5 \\
& Entropy       & 66.6 & 59.5 & 65.0 & 63.7 & 36.8 & 43.7 & 69.0 & 69.0 & 60.6 \\
& Margin        & \textbf{74.3} & 60.6 & \underline{69.5} & 68.1 & \underline{49.1} & \underline{53.9} & 74.4 & \textbf{73.0} & \underline{67.1} \\
& MPD-PAC       & 72.3 & 61.1 & 66.2 & 66.5 & 47.3 & 48.1 & 71.0 & 71.8 & 63.6 \\
& PC-Sampler    & 67.4 & 57.3 & 61.1 & 61.9 & 30.2 & 50.6 & \textbf{75.7} & 72.4 & 66.2 \\
& Info-Gain     & \underline{74.2} & \underline{62.7} & 68.3 & \underline{68.4} & 49.0 & 51.1 & 73.3 & 72.2 & 65.5 \\
\cmidrule(lr){2-11}
& \textbf{VIG-Sampler}  & \textbf{74.3} & \textbf{64.0} & \textbf{71.8} & \textbf{70.0} & \textbf{59.5} & \textbf{56.1} & \underline{75.6} & \underline{72.8} & \textbf{68.2} \\

\bottomrule
\end{tabular*}
\caption{
Experimental results of LLaDA-V with parallel decoding budgets $k \in \{2,4,8\}$.
The first and last Avg.\ columns report mean CIDEr and VQA scores, respectively.
The best scores are shown in \textbf{bold}, and the second-best scores are \underline{underlined}.
}
\label{tab:main_results_llada_v}
\end{table*}

\subsection{Overall Decoding Procedure of VIG-Sampler}
For completeness, we summarize the overall decoding procedure of VIG-Sampler in Algorithm~\ref{alg:vig_decoding}.

\begin{algorithm}[H]
\caption{Visual Information-Guided Sampler}
\label{alg:vig_decoding}
\begin{algorithmic}[1]

\REQUIRE Visual input $\mathbf{I}$, text prompt $\mathbf{x}$,
response length $N$, commit budget $k$, and
hyperparameters $\gamma,\lambda$
\ENSURE Decoded response $\mathbf{y}_T$

\STATE $\mathbf{y}_0 \gets [\mathrm{MASK}]^N$
\STATE $T \gets N/k$

\FOR{$t=0,\ldots,T-1$}

    \STATE $\mathcal{M}_t
    \gets
    \{i \mid y_t^i=[\mathrm{MASK}]\}$

    \STATE Run one forward pass conditioned on
    $(\mathbf{I},\mathbf{x},\mathbf{y}_t)$ to obtain
    $p_\theta^i(\cdot\mid\mathbf{y}_t)$ and $A$

    \FORALL{$i\in\mathcal{M}_t$}
        \STATE $\hat{y}_t^i
        \gets
        \arg\max_{v\in\mathcal{V}}
        p_\theta^i(v\mid\mathbf{y}_t)$

        \STATE $c_t^i
        \gets
        p_\theta^i(\hat{y}_t^i\mid\mathbf{y}_t)$

        \STATE $\mathbf{a}_t^i
        \gets
        A_{i,\mathcal{I}}$

        \STATE $m_t^i
        \gets
        \sum_{j\in\mathcal{I}} A_{i,j}$
    \ENDFOR

    \STATE $m_t^{\mathrm{med}}
    \gets
    \operatorname{median}_{i\in\mathcal{M}_t} m_t^i$

    \STATE $\bar{\mathbf{a}}_t
    \gets
    \dfrac{1}{|\mathcal{M}_t|}
    \sum_{j\in\mathcal{M}_t}\mathbf{a}_t^j$

    \FORALL{$i\in\mathcal{M}_t$}
        \STATE $r_t^i
        \gets
        c_t^i
        \left(
        \dfrac{m_t^i}{m_t^{\mathrm{med}}}
        \right)^\gamma$

        \STATE $\widetilde{\mathbf{a}}_t^i
        \gets
        \mathbf{a}_t^i-\bar{\mathbf{a}}_t$
    \ENDFOR

    \FORALL{$i,j\in\mathcal{M}_t,\ i\neq j$}
        \STATE $G_t^{i,j}
        \gets
        \max
        \left\{
        \left\langle
        \tilde{\mathbf{a}}_t^i,
        \tilde{\mathbf{a}}_t^j
        \right\rangle_{\mathrm{cos}},
        0
        \right\}$
    \ENDFOR

    \STATE $\mathcal{S}\gets\emptyset$

    \WHILE{$|\mathcal{S}|<k$}

        \FORALL{$i\in
        \mathcal{M}_t^{(\mathcal{S})}$}

            \IF{$\mathcal{S}=\emptyset$}
                \STATE $q_t^i\gets r_t^i$
            \ELSE
                \STATE $q_t^i
                \gets
                r_t^i
                -
                \dfrac{\lambda}{|\mathcal{S}|}
                \sum_{j\in\mathcal{S}}
                G_t^{i,j}$
            \ENDIF

        \ENDFOR

        \STATE $i^\star
        \gets
        \arg\max_{
        i\in\mathcal{M}_t^{(\mathcal{S})}}
        q_t^i$

        \STATE $\mathcal{S}
        \gets
        \mathcal{S}\cup\{i^\star\}$

    \ENDWHILE

    \STATE $\mathbf{y}_{t+1}\gets\mathbf{y}_t$

    \FORALL{$i\in\mathcal{S}$}
        \STATE $y_{t+1}^i\gets\hat{y}_t^i$
    \ENDFOR

\ENDFOR

\RETURN $\mathbf{y}_T$

\end{algorithmic}
\end{algorithm}

\section{Additional Experiments and Analysis}
\subsection{Generalization to a Different Model Family}
In Table~\ref{tab:main_results_llada_v}, we present additional experiments on LLaDA-V~\cite{llada-v} to evaluate the generalizability of VIG-Sampler to a different model family.

\subsection{Additional Ablation Study on Generation Length}
As shown in Table~\ref{tab:gen_length_ablation_k2}, we present additional comparisons between VIG-Sampler and Info-Gain across different generation lengths with $k=2$.
Consistent with the results at $k=4$, VIG-Sampler maintains strong overall performance across different generation lengths, demonstrating its robustness to the choice of $N$.

\begin{table}[t]
\centering
\begin{tabular}{c|c|cc}
\toprule
$N$ & Sampler & COCO Cap. & TextVQA \\
\midrule

\multirow{2}{*}{16}
& Info-Gain    & 87.2  & \textbf{58.0} \\
& VIG-Sampler  & \textbf{92.4}  & 57.5 \\
\cmidrule{1-4}

\multirow{2}{*}{32}
& Info-Gain    & 106.0 & 57.4 \\
& VIG-Sampler  & \textbf{106.2} & \textbf{58.7} \\
\cmidrule{1-4}

\multirow{2}{*}{48}
& Info-Gain    & \textbf{106.8} & 56.5 \\
& VIG-Sampler  & 104.2 & \textbf{59.4} \\
\cmidrule{1-4}

\multirow{2}{*}{64}
& Info-Gain    & 104.6 & 57.3 \\
& VIG-Sampler  & \textbf{105.9} & \textbf{59.0} \\
\bottomrule
\end{tabular}
\caption{Robustness to $N$ using LaViDa at $k=2$.}
\label{tab:gen_length_ablation_k2}
\end{table}

\subsection{Analysis of Inference Cost}

Table~\ref{tab:inference_cost} summarizes the inference cost of each sampling method in terms of peak GPU memory and wall-clock time.
VIG-Sampler shows no measurable increase in peak memory compared with the baselines, with all methods using 17.9 GB. 
Its wall-clock time also remains close to that of confidence-based sampling for both $k=2$ and $k=4$, while being substantially lower than that of Info-Gain. 
These results indicate that VIG-Sampler introduces only modest computational overhead despite incorporating visual guidance into token selection.

\begin{table}[t]
\centering
\setlength{\tabcolsep}{6pt}
\renewcommand{\arraystretch}{1.1}

\begin{tabular}{l|c|cc}
\toprule
\multirow{2}{*}{Sampler}
& \multirow{2}{*}{\shortstack{Peak mem.\\(GB)}}
& \multicolumn{2}{c}{Wall-clock time (s)} \\
\cmidrule(lr){3-4}
& & $k{=}2$ & $k{=}4$ \\
\midrule
Confidence  & 17.9 & 1.71 & 1.39 \\
Info-Gain   & 17.9 & 3.25 & 2.21 \\
VIG-Sampler & 17.9 & 1.85 & 1.44 \\
\bottomrule
\end{tabular}

\caption{Inference cost of each sampler.}
\label{tab:inference_cost}
\end{table}

\subsection{Additional Qualitative Case Studies and Analysis of VIG-Sampler}
\begin{figure*}[t!]
  \centering
  \includegraphics[width=\textwidth]{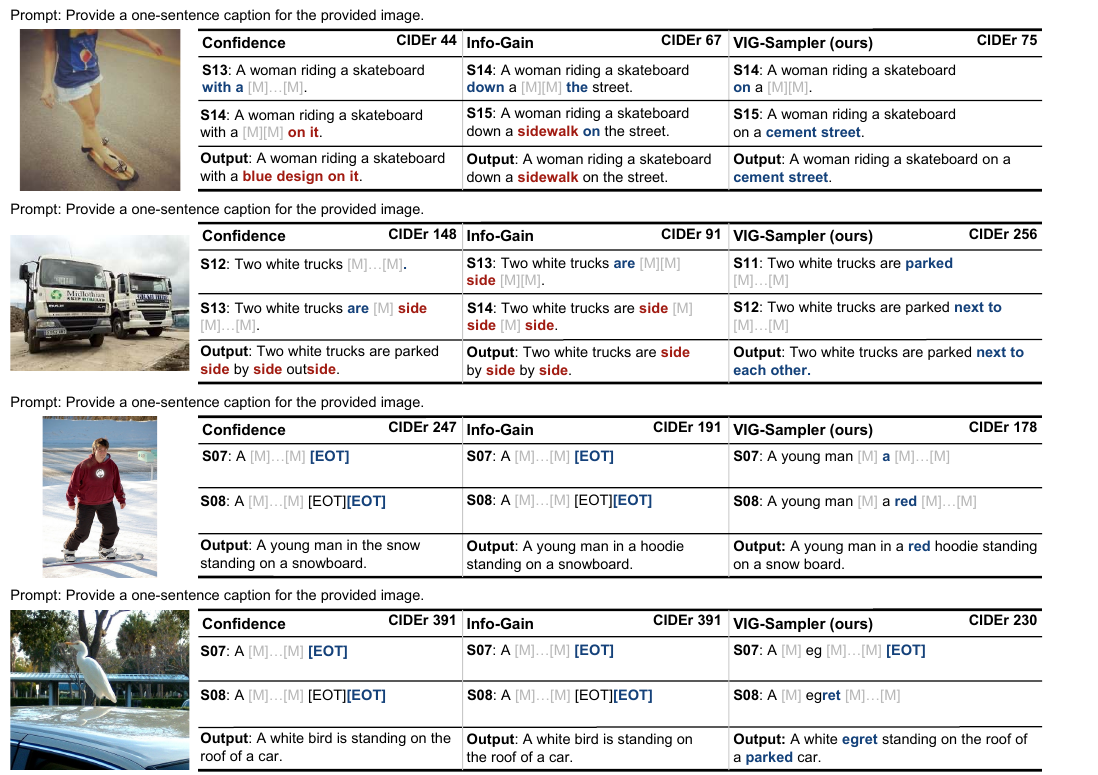}
\caption{Qualitative examples on COCO Caption using LaViDa at $k=2$. The top two rows illustrate decoding failures by conventional methods without explicit visual guidance, in which incorrect tokens are produced during generation. 
The bottom two rows present detailed captions that are visually well aligned with the images from a human perspective, despite receiving relatively low CIDEr scores.
\textbf{S\#} denotes the decoding step, while red and blue indicate incorrectly and correctly committed tokens, respectively.}
  \label{fig:qualitative_2}
\end{figure*}

In Figure~\ref{fig:qualitative_2}, we provide additional qualitative comparisons that highlight the effectiveness of VIG-Sampler.

\begin{figure*}[t!]
  \centering
  \includegraphics[width=\textwidth]{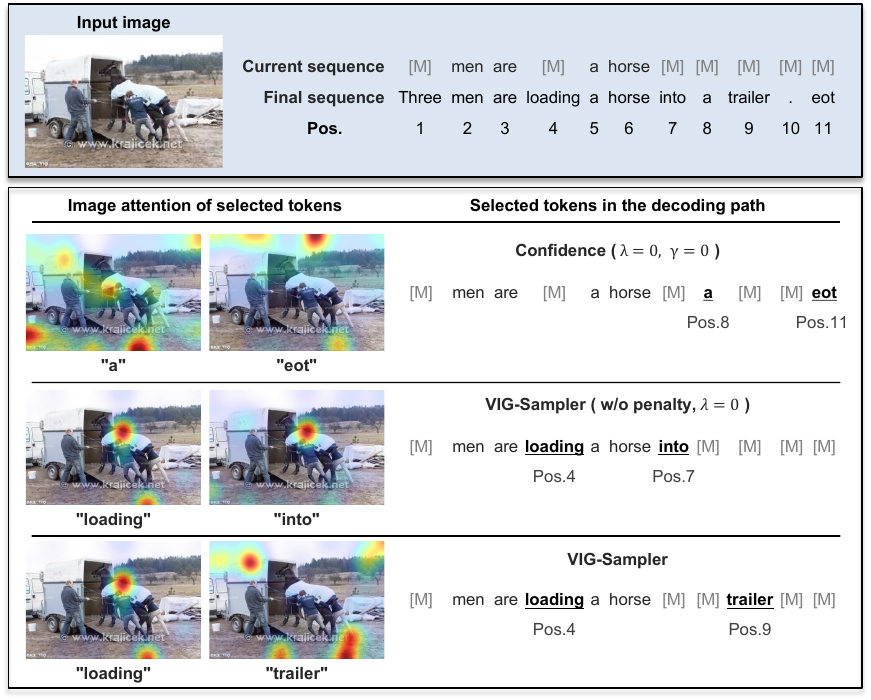}
    \caption{Qualitative analysis of token selection and image-attention patterns. 
    Confidence-based selection tends to choose uninformative tokens with diffuse attention, whereas VIG-Sampler selects answer-relevant tokens grounded in distinct image regions. 
    Without the penalty term, VIG-Sampler selects tokens with similar image-attention patterns, providing little additional information beyond one another.}
  \label{fig:qualitative_att}
\end{figure*}

\noindent We further report the image-attention patterns of the selected tokens in Figure~\ref{fig:qualitative_att}. 
In this example, the confidence-based method selects uninformative tokens with diffuse attention. 
Unlike the confidence-based method, VIG-Sampler selects tokens that are critical to generating the final answer and grounds them in distinct image regions.
Additionally, we investigate the effect of a penalty based on image-attention similarity on token selection during decoding.
Under the $\lambda=0$ ablation setting (i.e., with the penalty term disabled), the selected tokens attend to nearly identical image regions and provide little additional information beyond one another.
By contrast, in our default setting with the penalty term, the selected tokens attend to distinct image regions and faithfully capture different visual content.

\end{document}